# Some Critical and Ethical Perspectives on the Empirical Turn of AI Interpretability

Jean-Marie John-Mathews[a]

*Université Paris-Saclay, Univ Evry, IMT-BS, LITEM, 91025, Evry, France*

# Some Critical and Ethical Perspectives on the Empirical Turn of AI Interpretability

We consider two fundamental and related issues currently faced by Artificial Intelligence (AI) development: the lack of ethics and interpretability of AI decisions. Can interpretable AI decisions help to address ethics in AI? Using a randomized study, we experimentally show that the empirical and liberal turn of the production of explanations tends to select AI explanations with a low denunciatory power. Under certain conditions, interpretability tools are therefore not means but, paradoxically, obstacles to the production of ethical AI since they can give the illusion of being sensitive to ethical incidents. We also show that the denunciatory power of AI explanations is highly dependent on the context in which the explanation takes place, such as the gender or education level of the person to whom the explication is intended for. AI ethics tools are therefore sometimes too flexible and self-regulation through the liberal production of explanations do not seem to be enough to address ethical issues. We then propose two scenarios for the future development of ethical AI: more external regulation or more liberalization of AI explanations. These two opposite paths will play a major role on the future development of ethical AI.

**Introduction**

While Artificial Intelligence (AI) is increasingly used in Information Systems (IS), with efficient results (Little, 2004), it poses many problems because it produces decisions that are difficult to explain, which in turn create many challenges in terms of accountability, machine usability, trust, etc. A large and very active community of computer scientists, called XAI (Explainable Artificial Intelligence), have recently developed many analytics tools to make these AI systems interpretable.

However, XAI systems face a problem: the concept of AI interpretability is difficult to formalize independently of context (Miller, 2019). There is consensus today that explanations do not intrinsically emerge from decisions but always depend on the context in which the explanation takes place (Lipton, 2018). An algorithmic decision is explained differently, depending on the person to whom it is explained, the reason why the explanation is needed, the place and time of the explanation, the ergonomics of human-machine interaction, and so on.

To solve this context dependency problem, engineers and researchers increasingly tend to directly test different kinds of explanations on the end user of the AI algorithm, in order to establish the best one. Interpretability is then defined as a property of an algorithm that empirically reaches at best some end-user desiderata. There are several possible desiderata for an AI explanation: it must be actionable, understandable, increase satisfaction and confidence, and so on (Sokol and Flach, 2020). In the same way that Machine Learning integrates the decision context through a trial-and-error procedure that tests on a historical database, the production of AI explanations also becomes a trial-and-error procedure that tests several explanation devices according to well-defined end-user desiderata (Doshi-Velez and Kim, 2017; Miller et al., 2017; Poursabzi-Sangdeh et al., 2018). We will call this turning point the empirical or liberal turn of XAI.

We argue in this paper that this increasingly empirical approach to AI interpretability will run up against an important problem, namely AI ethics. As AI is increasingly criticized for social and ethical issues (McGregor, 2020), AI development is expected to integrate, in its process, a number of normative principles, such as fairness, privacy, etc. However, these normative principles are not supposed to depend on the context of explanation. For example, if an algorithmic decision is discriminatory, a good explanation of this decision should be able to accurately reveal the discriminatory aspect regardless of the context of explanation. A sexist algorithm is indeed expected to be interpreted as sexist regardless of the reviewer's level of education, the technical method of explanation, etc. Ethical principles have a certain universality and objectivity that should ideally not be altered in relation to the ways in which algorithms are examined[1]. We argue that the antagonism between context-dependent explanations and context-independent ethical principles is a major issue for the development of ethical artificial intelligence in the future.

What are the expected long-term impacts of the empirical turn of XAI on AI development and society? What are the limitations of XAI in addressing AI ethics-related issues? How to reconcile context-independent ethical principles and context-dependent empirical methods for the future development of AI?

To answer this question, we introduce an additional desideratum for explanation: denunciatory power. This desideratum is tested empirically in a randomized experimental study where respondents are confronted with an algorithmic ethical incident through an interpretation tool. Since the empirical turn of XAI empirically tests explanations according to predefined desiderata, this randomized study allows us to anticipate the evolution of XAI methods in the near future. While better transparency and interpretability is often considered as an answer to solve ethical issues generated by AI (Burrell, 2016), we empirically show that the aforementioned empirical trend in the development of AI explanations favors explanations that are blind to ethical incidents, which makes it difficult to set up safeguards for a more ethical AI. In other words, we show that self-regulation in AI ethics leads to select tools that paradoxically tend to dissimulate ethical incidents. Finally, we propose two possible scenarios of XAI development to address this limitation.

---

[1] Privacy or fair treatment are fundamental rights in various legal traditions.

# 1. Denunciatory power as a desideratum for empirical explanation

## *2.1. Interpretability methods for AI (XAI)*

Artificial Intelligence (AI) systems are increasingly used in a wide variety of sectors such as human resources, banking (Wang et al., 2018), health (Wang et al., 2020) and legal services, for tasks such as job candidate screening (Liem et al., 2018), medical diagnosis (Kononenko, 2001), judicial sentencing (Kleinberg et al., 2017), etc. These systems are often accurate and they significantly impact businesses and organizations (Little, 2004), yet they spawn many new challenges. One of the main challenges is the lack of interpretability of some ML algorithms. The reason is that some models are highly multi-dimensional, use non-linear transformations, and are fundamentally heuristic since they are not based on probabilistic theories that ensure their mathematical validity (Kraus and Feuerriegel, 2017; Mahmoudi et al., 2018). These models, such as Deep Learning, are often called black boxes or non-transparent algorithms in the ML literature (Lipton, 2016). It is crucial for the development of future AI to be able to generate explanations for its decisions, in order to guarantee user confidence, improve human-machine interaction or decision operability.

In this context, a very active research initiative called Explainable Artificial Intelligence (XAI) proposes many tools and methods to make algorithms interpretable (Mohseni et al., 2018). One of the main families of methods proposes post-hoc explanations of a decision resulting from a black-box algorithm (e.g. Shapley value or counterfactual explanations). At the same time, there is a contrasting method, consisting in not using black box algorithms and trying to improve the performance of intrinsically interpretable algorithms (transparent algorithm), such as linear model or tree-based models (Rudin, 2019). Figure 1 provides an example of a post-hoc explanation of a black-box AI decision (refusal of credit demand), giving Shapley importance values for 6 variables inside the AI model.

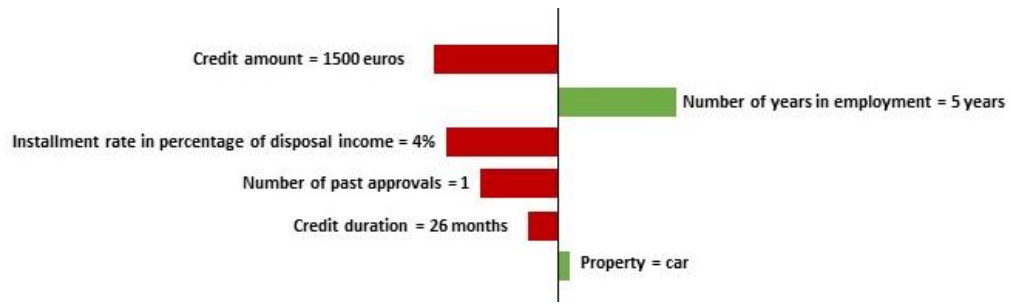

*Figure1: Variable importance (Shapley values) for 6 features involved in a AI model*

## 2.2. The empirical turn of XAI

The XAI community usually claims that interpretability is a difficult concept to define because it is too dependent on the context of explanation (Lipton, 2018). Interpretability is then considered as a latent property that cannot be explicitly formalized but whose consequences on the end-user can be tested empirically, depending on the context (Poursabzi-Sangdeh et al., 2018). Thus, a good explanation is the one that best answers experimentally to some contextual desiderata on the end-user (understanding, actionability, trust, fairness, etc.). We call these explanations obtained by testing according to end-user desiderata, empirical explanations. They no longer seek to be objective and context-independent, but instead adapt to context through pre-established objectives (desiderata) that are empirically tested (Doshi-Velez and Kim, 2017).

In a context where ethics plays an important role, we argue that it is essential to introduce an additional desideratum to test empirically: the capacity of the explanation to denounce an unethical algorithmic decision. In other words, we consider that a good explanation when addressing an unethical incident is one that allows the incident to be revealed to the end user. We refer to this desideratum as the denunciatory power of explanation. We argue in this paper that it is highly important to introduce this additional desideratum because it is rarely reached using the empirical approach of interpretability.

*2.3. How to measure the impact of the empirical trend in XAI?*

To support our thesis, we designed an experimental study where respondents were confronted with an algorithmic ethical incident through an interpretation device. To measure the explanation's capability to denounce, we measured what we call the end user's negative reception, using four variables: perception of the fairness of the decision; trust in the decision, enunciation of negative criticism about the decision; and willingness to challenge the decision.

**2. Data and Methods**

We designed a randomized experiment on 800 people with 2x4 scenarios, using a consumer panel collected over 2 weeks in June 2020 by Kantar. Each scenario was tested on a random sample of 100 individuals representative of the population, using statistical quotas for gender and age. We also verified that none of the 8 scenarios was statistically distributed differently in terms of education level and socio-professional category, compared to the total population. For each scenario, we asked the participants to imagine that they applied to a bank for a loan and that the bank was using an artificial intelligence algorithm to accept or refuse the application (see Appendix A). To build the AI algorithm we used the German Credit Scoring [39] database from the UC Irvine Machine Learning Repository as a training dataset for our ML algorithm. This dataset consisted of 1000 individual profiles, each with 20 features and categorized as either "good" or "bad" credit risks. From the 1000 individuals in the German Credit Scoring database, we built two Machine Learning algorithms to decide whether to accept or refuse the credit application: a logistic regression with integer coefficients (for the transparent explanation scenarios, see below) and a 2-layer feedforward neural network of 10 neurons in each layer (for the post-hoc explanation scenarios, see below). Using these two Machine Learning algorithms, we constructed 2 "ethical" configurations on 4 explanation configurations, making a total of 8 scenarios (2x4).

For each scenario, we started by introducing the case, then gave the input variables used as input by the AI algorithm, and informed the participant that his or her credit had been refused by the algorithm (see Appendix A).

*2.1 The two "ethical" configurations*

These two configurations corresponded respectively to cases where the algorithmic decision was "sexist" or not. To simulate the sexist algorithmic decision, we identified, within the German Credit scoring database, the case of a woman whose loan application had been refused by our algorithm and who would have been accepted if she had not been a woman[2]. Once this case was identified, we asked the respondents to imagine that they applied to a bank for a loan, and that they had the same characteristics as this woman identified in German Credit Scoring. We then build two scenarios with these characteristics: one where the gender variable was integrated into our learning model, and one where the gender variable was not integrated into the learning model. We refer to this first scenario as the "sexist decision" scenario because it simulates the decision of an algorithm that denies bank credit to a woman because of her sex. We refer to the second scenario as the "without incident" scenario because it does not take into account the sex variable in its decision[3]. By choosing a non-fictive case within the German Credit Scoring database, we are looking for a non-theoretical and realistic simulation of a sexist incident.

---

[2] We chose a woman in the dataset who maximized the difference between the probability of default and the probability of default if she had been a man.

*2.2 The four explanation configurations*

*2.2.1. No explanation scenarios*

In the first scenario, we did not provide any further explanation other than the aforementioned introduction[4] (Appendix A). This scenario was used as a baseline to measure the effects of the other scenarios.

*2.2.2. Transparent algorithm scenarios*

In this second configuration, we provided a points-based system using the integer coefficient of the logistic regression model to explain why the credit decision was negative. To do so, we assigned to each feature a certain number of points obtained from the coefficients of the logistic regression. If the sum of all the points exceeded a certain threshold, the bank loan was accepted (see Appendix C). Concerning the transparent scenario with the sexist decision, we mentioned that the number of points attributed to women was lower than the number of points attributed to men. (see Appendix B). The linearity of the logistic regression made it easy to represent the decision rule as a sum of points. That is why, in the XAI literature, logistic regression with few variables is considered as a transparent or intrinsically interpretable ML algorithm (Rudin, 2019). We refer to these two scenarios as the transparent algorithm scenarios.

---

[3] This consideration has limitations: even if the algorithm does not take into account the sex variable, it can have a differential treatment according to gender if the other variables approximate the sex variable. However, for our experiment, this is not a problem because the explanations given to the respondents (transparent explanation, Shapley, counterfactual) do not detect the differential treatment when the sex variable is not present in the model. See next section.

[4] Note that this introduction differs depending on whether or not the case is sexist. In fact, the sex variable is present in the introduction for scenarios with sexist decisions. Consequently, scenarios without explanations can give different answers, depending on the presence or not of the sexist decision.

*2.2.3. Post-hoc Shapley scenarios*

In the third configuration, we used the neural network instead, to produce the algorithmic decision. Contrary to the logistic regression, since the neural network could not be represented as a simple linear additive model, we provided a post-hoc explanation of the decision, by means of a feature relevance method on a local decision. To do so, we provided Shapley values, one of the most commonly used methods for post-hoc explanation of black-box models (Lundberg et al., 2017). In other words, we informed participants of the importance of each feature with respect to the negative decision made by the black-box algorithm (see Appendix D and E). We referred to these two scenarios (sexist and non-sexist decision) as the Shapley post-hoc explanation scenarios.

*2.1.4. Post-hoc counterfactual scenarios*

In this last configuration, we used the same black-box neural network to make the decision to refuse credit. However, instead of giving feature relevance for each variable, we gave counterfactual explanations ('What if' explanations) for the two actionable features, that is, credit duration and credit amount (Barredo Arrieta et al., 2020). To generate counterfactual explanations, we found, for both variables, the threshold at which the negative decision switched to a positive one. Applied to the two actionable features (i.e. features that could be changed or modified by the credit demander, such as credit amount or duration), this method gave two different counterfactual explanations that could help participants to know how to modify their application in order to change the algorithm output (see Appendix F)[5].

*2.3 Negative reception of the end user*

To measure the explanation's capability to denounce, we measured what we call the negative reception of the end user, using four variables: perception of the fairness of the decision; trust in the decision; emission of a negative criticism about the decision; and willingness to challenge the decision.

To do so, we asked the participants to rate their agreement with two perception desiderata and two reaction desiderata:

- Fairness perception of the algorithmic decision using a scale from "-2" (strongly disagree) to "2" (strongly agree).

- Trust perception of the algorithmic decision using a scale from "-2" (strongly disagree) to "2" (strongly agree).

- Free comments: we left a free field for respondents to comment freely on the algorithmic decision. From this free field, we counted the number of positive or neutral comments and the number of negative (critical) comments. In all, out of 800, 46% of respondents made a negative comment on the algorithmic decision and 54% made no comment. No respondent made a positive comment, which was not surprising since each scenario simulated a credit refusal. When the comment was negative, we counted whether it was an internal or external criticism. An internal criticism is one that challenges the way the algorithm is calculated, using the elements given in the explanation of the decision (e.g. "The amount of the repayment is largely compatible with my income; "I've been working for 5 years and I've already been granted a loan in the past"). The external criticism more radically questions the algorithm without mentioning precisely how it works (e.g. "A banking decision should be more humane"; "it is an arbitrary decision"). Among the negative comments, we found 74% internal criticisms and 26% external criticisms.

---

[5] For the counterfactual configuration, the scenarios with and without the sexist decision provided the same explanation since gender is not an actionable variable. However, these two scenarios differed because of their different introduction (See Appendix A).

- Wish to contest the algorithmic decision using a scale from "-2" (strongly disagree) to "2" (strongly agree).

We began the interview by asking the participants about some socio-demographic variables such as their level of education (graduation from higher education), gender, etc. To ensure that they answered the questions carefully, we asked them attention questions about their understanding of the statement, which allowed us to filter out 10% of the participants that did not read the statements carefully.

## 3. Results

We show in this section that the empirical turn of interpretation tends to select the modes of explanation with the weakest denunciatory power in the presence of an ethical incident.

### *3.1 Denunciatory power differs depending on modes of explanation*

Table 1 describes the denunciatory power of each mode of explanation. By comparing the negative reception between ethical situations using Student's tests, we see that only the Shapley explanations have a significant denunciatory power (Table 1). Conversely, the counterfactual explanations have the lowest denunciatory power because they are not significant for any of the dimensions of negative reception. The denunciatory power of the explanations is therefore dependent on the modes of technical explanation.

|  |  |  | \multicolumn{8}{c}{**RECEPTION**} |
|---|---|---|---|---|---|---|---|---|---|---|
|  |  |  | \multicolumn{2}{c}{**Fairness perception**} | \multicolumn{2}{c}{**Trust perception**} | \multicolumn{2}{c}{**Negative comments rate**} | \multicolumn{2}{c}{**Claim rate**} |
|  |  |  | With Incident | Without Incident | With Incident | Without Incident | With Incident | Without Incident | With Incident | Without Incident |
| TECHNICAL INTERPRETABILITY | TRANSPARENT |  | 3.4 | 3.7 | 3.7 | 3.7 | **51%** | **35%** | 55% | 39% |
|  |  | pvalue | *(0.2)* |  | *(1)* |  | ***(0.02)*** |  | *(0.3)* |  |
|  | POST-HOC SHAPLEY |  | **3.3** | **3.7** | **3.3** | **3.7** | 55% | 44% | 66% | 43% |
|  |  | pvalue | ***(0.07)*** |  | ***(0.05)*** |  | *(0.1)* |  | *(0.15)* |  |
|  | POST-HOC COUNTER-FACTUAL |  | *3.5* | *3.5* | *3.7* | *3.6* | *38%* | *39%* | *37%* | *20%* |
|  |  | pvalue | *(0.7)* |  | *(0.7)* |  | *(0.9)* |  | *(0.3)* |  |

*Table 1: Denunciatory power depending on explanation modes (t-test with H0: means equality with the baseline scenario where decisions are not sexist)*

**3.2 Empirical explanations tend to select the explanation that has the weakest denunciatory power**

In the empirical approach, for AI interpretability, the mode of explanation that is chosen is the one that best responds to the desiderata. Since negative reception of algorithmic decisions is inversely correlated to the economic value of the algorithm, the algorithm owner will choose the mode of explanation that significantly decreases the negative reception compared to the situation without explanation. Table 2 shows that only the counterfactual explanation, which is the mode with the lowest denunciatory power (see above), presents significant differences compared to the scenario without explanation. It will then be selected by the AI designer to provide an explanation to end users since it meets, at best, the pre-determined desiderata for interpretability (empirical turn of explanation building).

|  |  |  | RECEPTION | | | |
|---|---|---|---|---|---|---|
|  |  |  | **Fairness perception** | **Trust perception** | **Negative Comments rate** | **Claim rate** |
| **TECHNICAL INTERPRETABILITY** | **NO EXPLANATION (BASELINE)** |  | 3.48 | 3.38 | 55% | 58% |
|  | **TRANSPARENT** |  | 3.45 | **3.7** | 51% | 55% |
|  |  | pvalue | *(0.9)* | ***(0.1)*** | *(0.6)* | *(0.9)* |
|  | **POST-HOC SHAPLEY** |  | 3.3 | 3.3 | 55% | 66% |
|  |  | pvalue | *(0.3)* | *(0.8)* | *(1)* | *(0.6)* |
|  | **POST-HOC COUNTERFACTUAL** |  | 3.45 | **3.7** | **38%** | **37%** |
|  |  | pvalue | *(0.9)* | ***(0.1)*** | ***(0.016)*** | ***(0.1)*** |

*Table 2: Selection of best explanation modes when AI incident is present (t-test with H0: means equality with the baseline scenario where there is no explanation). Population: 400 individuals corresponding to situations with sexist incidents.*

The empirical design of the explanation tends to value the explanation with the lowest denunciatory power when addressing an AI incident. This is because the empirical approach seeks explanations that minimize the number of negative feedbacks of the AI decision, while denunciatory power is measured precisely through the negative feedbacks. In this situation with contradictory objectives, denunciatory power is likely to be neglected in the development of future artificial intelligence tools, to minimize the number of negative feedbacks. This is a direct consequence of the empirical explanation trend in AI, where explanations are empirically tested using desiderata.

This result is consistent with some recent theoretical works on AI Ethics claiming that AI designers may choose algorithms that align with what is for them the most convenient epistemological understanding of an ethical principle, rather than the one that aligns with society's preferred understanding (Krishnan, 2020). Other authors point out that stated motivations for addressing ethics might not reflect actual motivation (Schiff et al., 2019) and warn against "fair-washing" (Aïvodji et al., 2019). We show empirically, using an experimental setup, that it is possible for AI designers to propose modes of technical explanation that lower the level of criticism and avoid revealing unethical situations. Far from being a solution to the ethical incidents of AI, explanation techniques can be hijacked by manufacturers in their interests, to the detriment of ethics. But we show that this phenomenon is not necessarily intentional on the part of the designer: by empirically selecting explanations with respect to desiderata that reduce criticism, the manufacturer indirectly creates the structural conditions for masking ethical incidents.

**4. The denunciatory power of explanations may vary according to contexts**

Since XAI empirical methods tend to choose explanations with the lowest denunciatory power, an empirical solution to the problem would be to add the denunciatory power of explanations as a desideratum to be empirically tested. However, we see in this section that this approach also has limitations.

Let us now suppose that the designer can add the denunciatory power as a desideratum to be tested and that it chooses the mode of explanation with the greatest denunciatory power. In this case, for this last section, counterfactual explanations are filtered out so that only Shapley explanations and transparent algorithms are kept.

We are now interested in the variability of the denunciatory power of an explanation according to the context of the explanation. Let us examine the variability of the denunciatory power according to the profile of the examiner: sex and level of education.

*4.1. Denunciatory power depending on sex*

The Student's tests being more significant on the first line, Table 3 shows that the denunciatory power of the explanation is stronger for women than for men. This result can be interpreted. As the simulated accident is a sexist decision, we can expect women to be more sensitive than men on the question of sexism: the sexist incident increases the negative response in women.

|        |        |        | RECEPTION              |                    |                   |                    |                             |                    |                |                    |
|--------|--------|--------|------------------------|--------------------|-------------------|--------------------|-----------------------------|--------------------|----------------|--------------------|
|        |        |        | Fairness perception    |                    | Trust perception  |                    | Negative Comments rate      |                    | Claim rate     |                    |
|        |        |        | With Incident          | Without Incident   | With Incident     | Without Incident   | With Incident               | Without Incident   | With Incident  | Without Incident   |
| GENDER | FEMALE |        | **3.2**                | **3.7**            | 3.5               | 3.7                | **49%**                     | **34%**            | **61%**        | **33%**            |
|        |        | pvalue | *(0.02)*               |                    | *(0.3)*           |                    | *(0.03)*                    |                    | *(0.07)*       |                    |
|        | MALE   |        | 3.5                    | 3.7                | 3.6               | 3.8                | 56%                         | 46%                | 60%            | 49%                |
|        |        | pvalue | *(0.5)*                |                    | *(0.5)*           |                    | *(0.1)*                     |                    | *(0.5)*        |                    |

*Table 3: Denunciatory power depending on gender (t-test with H0: means equality with the baseline scenario where decisions are not sexist)*

## 4.2. Denunciatory power depending on education level

|           |                  |        | RECEPTION           |                  |                  |                  |                        |                  |                |                  |
|-----------|------------------|--------|---------------------|------------------|------------------|------------------|------------------------|------------------|----------------|------------------|
|           |                  |        | Fairness perception |                  | Trust perception |                  | Negative Comments rate |                  | Claim rate     |                  |
|           |                  |        | With Incident       | Without Incident | With Incident    | Without Incident | With Incident          | Without Incident | With Incident  | Without Incident |
| EDUCATION | HIGHER EDUCATION |        | **3.2**             | **3.7**          | **3.5**          | **3.8**          | **62%**                | **46%**          | **78%**        | **47%**          |
|           |                  | pvalue | *(0.04)*            |                  | *(0.1)*          |                  | *(0.01)*               |                  | *(0.03)*       |                  |
|           | LOWER EDUCATION  |        | 3.5                 | 3.7              | 3.6              | 3.6              | 40%                    | 31%              | 36%            | 32%              |
|           |                  | pvalue | *(0.4)*             |                  | *(1)*            |                  | *(0.2)*                |                  | *(0.8)*        |                  |

*Table 4: Denunciatory power depending on education level (t-test with H0: means equality with the baseline scenario where decisions are not sexist)*

Likewise, the denunciation power of explanations to those who have a higher level of education is stronger than to those who have not. However, this does not necessarily mean that those with higher education are more sensitive to sexism. In fact, we can see that those who have not had higher education tend to make external criticisms more often (43%) than those who have had higher education (17%). Since the criticism of sexism is part of an internal criticism, it becomes more frequent for those with higher education. It would therefore seem that the denunciatory power depends on the ability to make internal criticisms of a decision, which is itself correlated to education level.

*4.3. Discussion about the variability of denunciatory power of explanation*

We see that ethical issues in AI cannot be addressed by relying solely on empirical explicability and denunciatory power as desiderata of explanations. An explanation cannot by itself reveal an ethical incident: it still depends on the examiner, his or her level of education, gender, etc. This is problematic because the least endowed and the least concerned tend to detect ethical incidents less. If ethical incidents create inequalities, then the inequalities are likely to increase with the empirical approach to explanation since it harms people's agency to denounce injustice. Once again, the increase of inequalities is not necessarily an intentional fact on the part of the manufacturer but an indirect consequence of a liberal or empirical approach to interpretability in AI.

Yet, contrary to the desiderata of classical interpretability, which are context-specific, the denunciatory power of an ethical incident must be independent of the contexts of explanation. Otherwise, it is not possible to guarantee the ethical property of an algorithm if the provided explanation revealed the ethical incident only for certain people, in certain conditions. It is essential to guarantee equal capabilities to denounce between individuals in order not to increase existing inequalities. Ethical principles have a certain universality and objectivity that should not be altered according to the ways in which algorithms are examined.

## 5. Two scenarios for the development of XAI that reconciles context-independent ethical principles and context-dependent empirical methods

In a recent paper, (Morley et al., 2021) claim that AI Ethics "tools and methods are either too flexible (and thus vulnerable to ethics washing) or too strict (unresponsive to context)". In the same lines, how to reconcile context-independent ethical principles and context-dependent empirical methods for the future development of AI? We propose two scenarios that address this antagonism.

### *5.1. Un-liberalization of AI explanations*

By selecting explanations with the least negative feedback from users, self-regulation in AI tends to paradoxically eliminate methods of explanation with a strong denunciatory power. As a consequence, in this scenario, AI ethics cannot be solved by relying on explanations to denounce suspicious behaviors and individual behavior from end-user. It is therefore necessary to separate the concept of interpretability from user feedback. In this scenario, we need to define the concept of interpretability, so that it applies context-independently. Once it is formalized, experts or auditors can examine the ethics of algorithms independently of the user feedback. In this scenario, we suppose that context-independent ethical principles cannot be addressed by the production of context-dependent explanations. In other words, the ethical problem requires the empirical turn to turn back the clock, that is, to un-liberalize the process of making explanations. We need regulation to stabilize a definition of algorithmic interpretability and test it so that non-ethical behavior can be detected independently.

*5.2. Liberalization of AI explanations*

Another contrasting scenario would be to address the problems induced by liberal explanations by further liberalizing the process of making explanations. To do so, it is necessary to integrate the limits of the liberal turn as desiderata into the conception of new explanations. In this case, the denunciatory power of an explanation becomes a desideratum in the very construction of the explanations. To test the ability to denounce, the AI designer needs to reproduce our randomized experimental study, so to simulate an ethical incident and compare the user's reception with and without an ethical incident. Since the denunciatory power of explanations depends on the context of explanation, it will also be necessary to test the ability to denounce in as many contexts as possible (e.g. with testers with different socio-demographic profiles), to ensure that the explanation is valid according to the predefined contextual desiderata. The multiplication of testing scenarios is the price to pay for solving ethics through interpretability of algorithmic and avoid strong regulation.

**Conclusion**

In this paper, we consider two fundamental issues currently faced by AI development: the lack of ethics and the opacity of AI decisions. These two issues, which are at the origin of many research initiatives, seem to be closely linked. Is the interpretability of AI a means to achieve ethical AI? Being able to interpret the causes of an AI decision can indeed help to determine whether the decision is ethical or not. Interpretability then becomes a tool to denounce possible ethical incidents, which is what we have called the ability to denounce an AI explanation.

The community of researchers who produce AI explanations is currently experiencing an empirical turning point: interpretability is increasingly defined as a latent variable that is the cause of some testable desiderata on the end user. This is what we have called the empirical turn of XAI.

In this paper, we experimentally show that the empirical turn of the production of explanations will tend to select AI explanations with a low denunciatory power. While the empirical approach produces explanations that minimize the number of negative feedbacks from the end user, denunciatory power rather seeks situations that increase negative feedbacks. Faced with these contradictory objectives, we argue that denunciatory power is likely to be neglected in the development of future artificial intelligence devices in favor of minimizing the number of negative feedbacks.

Interpretability tools are therefore not means but, paradoxically, obstacles to the production of ethical AI since they can give the illusion of being sensitive to ethical incidents. This trend will generate risks, the management of which will represent a major challenge for the future development of artificial intelligence.

We propose two scenarios to meet this challenge. First, we can address the empirical turn of AI explanation by basing explanatory mechanisms on norms, standards, and legislation established by institutions or ethical committees, and not on individual end-user behavior. Second, we can continue along the path of the empirical turn by considering the denunciatory power as a desideratum to be tested by the designer on end-users. In this case, care must be taken to always multiply testing contexts in order to capture a maximum of different contexts and to test the equality of individuals' ability to denounce ethical incidents.

These two opposing scenarios actually present two very different visions of ethics: in the first case, ethics is a matter of universal principles, laws and norms applicable to all, whereas in the second case, ethics is a matter of individual preference, transparency and individual capability to denounce.

# Acknowledgments

# Appendices

## Appendix A: Introduction given to participants for each scenario

You apply for a loan from a bank. This bank uses an automatic artificial intelligence algorithm to decide whether to accept or refuse your application. The automatic algorithm uses the following 7 pieces of information about you to decide whether or not to grant you the loan:

1. Credit amount: €1500. This is the amount of the loan you are applying for.

2. Credit duration: 26 months. This is the time duration to pay back the loan.

3. Instalment rate in percentage of disposable income: 4%. This is the proportion of your monthly in-come that will be spent to pay back the loan.

4. Number of years in employment: 5 years. This is the number of years you have been employed.

5. Gender: Female <u>*(this information is not displayed for the four scenarios without sexist incidents)*</u>

6. Property: Car. These are the assets you own.

7. Past loan approvals: 1. This is the number of credit application approvals you have already obtained with the same bank in the past. In your case, you have therefore obtained only one bank credit approval with this bank in the past.

Your application is processed by an artificial intelligence algorithm. This algorithm decides whether or not to grant you the loan, based on these 6 variables and on past credit data from other bank customers.

Based on the information you provided, the automatic algorithm decides not to grant you the loan. Your application for bank credit is therefore refused.

## Appendix B: Transparent algorithm scenario – with sexist incident

To process your credit application, the automatic algorithm uses a points-based system. If the total number of points is greater than 210, then the algorithm accepts the credit application. Otherwise, the credit is refused. Below is a detailed explanation of how points are awarded:

1. ***Credit amount:***

- *Less than 1000 euros: 39 points*
- *Between 1000 and 2000 euros: 44 points*
- *Between 2000 and 3000: 47 points*
- *Between 3000 and 4000: 51 points*
- *Between 4000 and 5000: 37 points*
- *More than 5000: 42 points*

2. ***Credit duration:***

- *Less than 6 months: 37 points*
- *Between 6 and 12 months: 19 points*
- *Between 12 and 18 months: 10 points*
- *Between 18 and 24 months: 7 points*
- *Between 24 and 30 months: 9 points*
- *Between 30 and 36 months: 6 points*
- *36 months and over: 0 point*

3. ***Instalment rate in percentage of disposal income:***

- *1%: 37 points*
- *2%: 36 points*
- *3%: 39 points*
- *4%: 35 points*

4. ***Number of years in employment:***

- *Without work: 37 points*
- *Less than 1 year: 32 points*
- *From 1 to 4 years: 37 points*
- *From 4 to 7 years: 41 points*
- *Over 7 years: 41 points*

5. ***Gender:***

- *Male: 39 points*
- *Female: 37 points*

6. ***Property:***

- Life insurance: 43 points
- Real estate: 48 points
- Car: 44 points
- No assest: 37 points

7. **Number of credit application approvals in the past with the same bank:**

- More than two credit approvals in the past with the same bank: 37 points
- At most one credit approval in the past with the same bank: 33 points

Given your situation, you scored a total of 43 + 9 + 36 + 42 + 38 + 33 = 201 points.

Since your total number of points is less than 210 points, your credit application is refused.

## *Appendix C: Transparent algorithm scenario – without sexist decision*

To process your credit application, the automatic algorithm uses a points-based system. If the total number of points is greater than 210, then the algorithm accepts the credit application. Otherwise, the credit is refused. Below is a detailed explanation of how points are awarded:

1. **Credit amount:**

- Less than 1000 euros: 39 points
- Between 1000 and 2000 euros: 43 points
- Between 2000 and 3000: 47 points
- Between 3000 and 4000: 52 points
- Between 4000 and 5000: 37 points
- More than 5000: 42 points

2. **Credit duration:**

- Less than 6 months: 37 points
- Between 6 and 12 months: 19 points
- Between 12 and 18 months: 10 points
- Between 18 and 24 months: 7 points
- Between 24 and 30 months: 9 points
- Between 30 and 36 months: 5 points
- 36 months and over: 0 point

3. **Instalment rate in percentage of disposal income:**

- 1%: 37 points
- 2%: 36 points
- 3%: 39 points
- 4%: 36 points

4. **Number of years in employment:**

- Without work: 37 points
- Less than 1 year: 31 points
- From 1 to 4 years: 37 points
- From 4 to 7 years: 42 points
- Over 7 years: 41 points

5. **Property:**

- Life insurance: 37 points
- Real estate: 48 points
- Car: 38 points
- No asset: 37 points

6. **Number of credit application approvals in the past with the same bank:**

- More than two credit approvals in the past with the same bank: 37 points
- At most one credit approval in the past with the same bank: 33 points

Given your situation, you scored a total of 43 + 9 + 36 + 42 + 38 + 33 = 201 points.

Since your total number of points is less than 210 points, your credit application is refused.

## *Appendix D: Shapley Post-hoc explanation scenario – with sexist decision*

*The graph below gives the variables that impacted the algorithmic decision. A green bar indicates that the variable had a positive impact on your credit application, i.e. the variable increases the chances of approval of your application. On the contrary, a red bar indicates that the variable had a negative impact on your credit application, i.e. the variable decreases the chances of approval of your application. Finally, the bigger the size of the bars, the greater the influence of the variable on the processing of your application.*

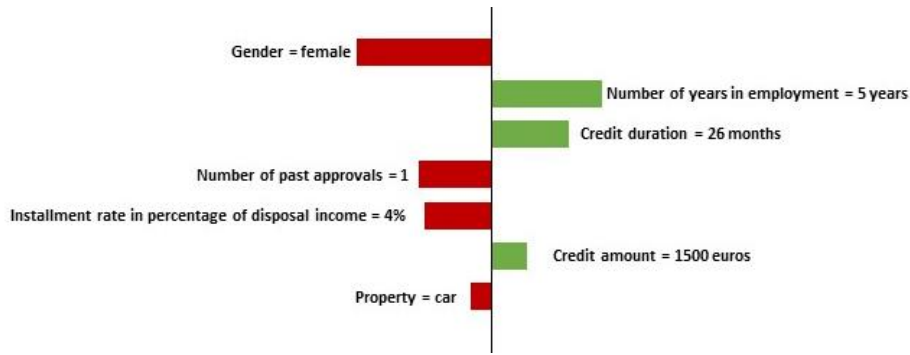

*Variable importance for the refusal decision*

## Appendix E: Scenario Shapley Post-hoc explanation – Without sexist decision

*The graph below gives the variables that impacted the algorithmic decision. A green bar indicates that the variable had a positive impact on your credit application, i.e. the variable increases the chances of approval of your application. On the contrary, a red bar indicates that the variable had a negative impact on your credit application, i.e. the variable decreases the chances of approval of your application. Finally, the bigger the size of the bars, the greater the influence of the variable on the processing of your application.*

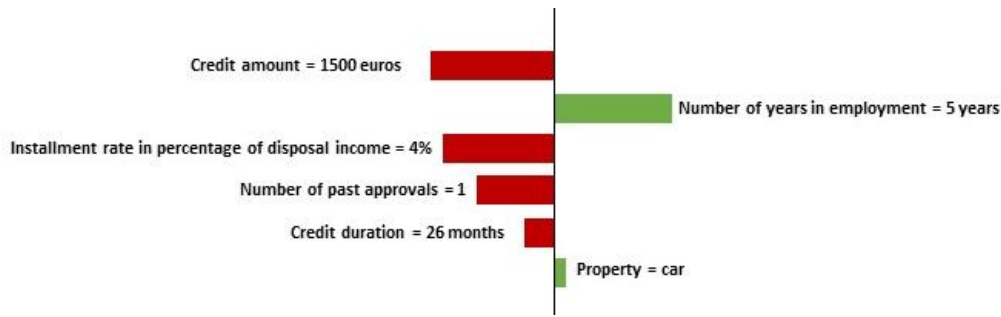

*Variable importance for the refusal decision*

## Appendix F: Counterfactual post-hoc explanation scenario – without and with sexist decision

*- If the credit amount was between 3000 and 4000 euros, your credit application would have been accepted by the algorithm. As a reminder, your credit amount is currently 1500 euros.*

*- If the credit duration was reduced to less than 12 months, your credit request would have been accepted by the algorithm. As a reminder, your credit duration is currently 26 months.*